\documentclass[11pt,a4paper]{article}
\usepackage[hyperref]{acl2020}
\usepackage{times}
\usepackage{latexsym}
\usepackage{amsmath}
\usepackage{amssymb}
\usepackage{hyperref}

\usepackage{microtype}

\usepackage{graphicx}
\usepackage{multirow}

\aclfinalcopy 
\def\aclpaperid{1033} 

\definecolor{ao(english)}{rgb}{0.0, 0.5, 0.0}
\definecolor{orange(ryb)}{rgb}{0.98, 0.6, 0.01}

\title{Variational Neural Machine Translation with Normalizing Flows}

\author{Hendra Setiawan \qquad Matthias Sperber \qquad Udhay Nallasamy \qquad Matthias Paulik \\
  Apple\\
  {\tt \{ hendra,sperber,udhay,mpaulik \}@apple.com} }

\date{}

\begin{document}
\maketitle

\begin{abstract}

Variational Neural Machine Translation (VNMT) is an attractive framework for modeling the generation of target translations, conditioned not only on the source sentence but also on some latent random variables. The latent variable modeling may introduce useful statistical dependencies that can improve translation accuracy. Unfortunately, learning informative latent variables is non-trivial, as the latent space can be prohibitively large, and the latent codes are prone to be ignored by many translation models at training time. Previous works impose strong assumptions on the distribution of the latent code and limit the choice of the NMT architecture. In this paper, we propose to apply the VNMT framework  to the state-of-the-art Transformer and introduce a more flexible approximate posterior based on normalizing flows. We demonstrate the efficacy of our proposal under both in-domain and out-of-domain conditions, significantly outperforming strong baselines. 

\end{abstract}

\section{Introduction}

Translation is inherently ambiguous. For a given source sentence, there can be multiple plausible translations due to the author's stylistic preference, domain, and other factors. On the one hand, the introduction of neural machine translation (NMT) has significantly advanced the field \cite{bahdanau+al-2014-nmt}, continually producing state-of-the-art translation accuracy. On the other hand, the existing framework provides no explicit mechanisms to account for translation ambiguity. 

Recently, there has been a growing interest in latent-variable NMT (LV-NMT) that seeks to incorporate latent random variables into NMT to account for the ambiguities mentioned above. For instance, \newcite{zhang-etal-2016-variational-neural} incorporated latent codes to capture underlying global semantics of source sentences into NMT, while \newcite{SuWXLHZ18} proposed fine-grained latent codes at the word level. The learned codes, while not straightforward to analyze linguistically, are shown empirically to improve accuracy. Nevertheless, the introduction of latent random variables complicates the parameter estimation of these models, as it now involves intractable inference. In practice, prior work resorted to imposing strong assumptions on the latent code distribution, potentially compromising accuracy.

In this paper, we focus on improving Variational NMT (VNMT)  \cite{zhang-etal-2016-variational-neural}: a family of LV-NMT models that relies on the amortized variational method \cite{DBLP:journals/corr/KingmaW13} for inference. Our contributions are twofold. (1) We employ variational distributions based on normalizing flows \cite{rezende15}, instead of uni-modal Gaussian. Normalizing flows can yield complex distributions that may better match the latent code's true posterior. (2) We employ the Transformer architecture \cite{transformer}, including \textit{Transformer-Big}, as our VNMT's generator network. We observed that the generator networks of most VNMT models belong to the RNN family that are relatively less powerful as a translation model than the Transformer. 

We demonstrate the efficacy of our proposal  on the German-English IWSLT'14 and English-German WMT'18 tasks, giving considerable improvements over strong non-latent Transformer baselines, and moderate improvements over Gaussian models. We further show that gains generalize to an out-of-domain condition and a simulated bimodal data condition. 

\section{VNMT with Normalizing Flows}

\noindent \textbf{Background} \quad Let $\boldsymbol{x}$ and $\boldsymbol{y}$ be a source sentence and its translation, drawn from a corpus $\mathcal{D}$. Our model seeks to find parameters $\theta$ that maximize the marginal of a latent-variable model $p_\theta(\boldsymbol{y},Z\mid \boldsymbol{x})$ where $Z \in \mathbb{R}^D$ is a sentence-level latent code similar to \cite{zhang-etal-2016-variational-neural}. VNMT models sidestep the marginalization by introducing variational distributions and seek to minimize this function (i.e., the Evidence Lower Bound or ELBO):
\begin{align}
\sum_{(\boldsymbol{x},\boldsymbol{y})\in\mathcal{D}} \mathbb{E}_{q(Z\mid \boldsymbol{x}, \boldsymbol{y})} \left[\log p_\theta(\boldsymbol{y}\mid\boldsymbol{x},Z) \right] \nonumber
\\
- \text{KL}\left( q(Z\mid \boldsymbol{x}, \boldsymbol{y}) \mid\mid p(Z\mid \boldsymbol{x})\right), \label{eq_elbo1}
\end{align}
\noindent where $q(Z\mid \boldsymbol{x}, \boldsymbol{y})$, $p(Z\mid \boldsymbol{x})$ are the variational posterior and prior distribution of the latent codes, while $p(\boldsymbol{y}\mid\boldsymbol{x},Z)$ is a \textit{generator} that models the generation of the translation conditioned on the latent code\footnote{In VAE terms, the posterior and prior distributions are referred to as the encoders, while the generator is referred to as the decoder. As these terms have other specific meaning in NMT, we avoid to use them in this paper.}. The ELBO is improved when the model learns a posterior distribution of latent codes that minimizes the reconstruction loss (the first term) while incurring a smaller amount of KL divergence penalty between the variational posterior and the prior (the second term). 

The majority of VNMT models design their variational distributions to model unimodal distribution via isotropic Gaussians with diagonal covariance, which is the simplest form of prior and approximate posterior distribution. This assumption is computationally convenient because it permits a closed-form solution for computing the KL term and facilitates end-to-end gradient-based optimization via the re-parametrization trick  \cite{rezende15}. However, such a simple distribution may not  be expressive enough to approximate the true posterior distribution, which could be non-Gaussian, resulting in a loose gap between the ELBO and the true marginal likelihood. Therefore, we propose to employ more flexible posterior distributions in our VNMT model, while keeping the prior a Gaussian.
\\\\
\noindent \textbf{Normalizing Flows-based Posterior} \quad \newcite{rezende15} proposed Normalizing Flows (NF) as a way to introduce a more flexible posterior to Variational Autoencoder (VAE). The basic idea is to draw a sample, $Z_0$, from a simple (e.g., Gaussian) probability distribution and to apply $K$ invertible parametric transformation functions $(f_k)$ called \textit{flows} to transform the sample. The final latent code is given by $Z_K = f_K ( ... f_2 ( f_1(Z_0)) ... )$ whose probability density function, $q_\lambda(Z_K \mid \boldsymbol{x}, \boldsymbol{y})$, is defined via the change of variable theorem as follows: 
\begin{align*}
q_0(Z_0 \mid \boldsymbol{x}, \boldsymbol{y}) \prod_{k=1}^K \left|\det\frac{\partial f_k(Z_{k-1}; \boldsymbol{\lambda}_k(\boldsymbol{x},\boldsymbol{y}))}{\partial Z_{k-1}}\right|^{-1},
\end{align*}
where $\lambda_k$ refers to the parameters of the $k$-th flow with $\lambda_0$ corresponds to the parameters of a base distribution. In practice, we can only consider transformations, whose determinants of Jacobians (the second term) are invertible and computationally tractable. 

For our model, we consider several NFs, namely \textit{planar flows} \cite{rezende15}, \textit{Sylvester flows} \cite{vdberg2018sylvester} and \textit{affine coupling layer} \cite{realnvp45819}, which have been successfully applied in computer vision tasks.

Planar flows (PF) applies this function: 
\begin{align*}
f_k(Z;\lambda_k(\boldsymbol{x},\boldsymbol{y})) = Z + \mathbf{u} \cdot \mathrm{tanh}(\mathbf{w}^TZ + \mathbf{b}),
\end{align*}
where $\boldsymbol{\lambda}_k = \{\mathbf{u}, \mathbf{w} \in \mathbb{R}^D$, $\mathbf{b} \in \mathbb{R}\}$. Planar flows perform contraction or expansion to the direction perpendicular to the  $(\mathbf{w}^TZ+ \mathbf{b})$ hyperplane.

Sylvester flows (SF) applies this function:
\begin{align*}
f_k(Z;\lambda_k(\boldsymbol{x},\boldsymbol{y})) = Z + \mathbf{A} \cdot \mathrm{tanh}(\mathbf{B}Z + \mathbf{b}),
\end{align*}
where $\boldsymbol{\lambda}_k = \{\mathbf{A}, \mathbf{B} \in \mathbb{R}^{M \times D}$, $\mathbf{b} \in \mathbb{R}^M\}$ and $M$ is the number of hidden units. Planar flows are a special case of Sylvester flows where $M=1$.  In our experiments, we consider the orthogonal Sylvester flows \cite{vdberg2018sylvester}, whose parameters are matrices with $M$ orthogonal columns. 

Meanwhile, the affine coupling layer (CL) first splits $Z$ into $Z^{d_1}, Z^{d_2} \in \mathbb{R}^{D/2}$ and applies the following function:
\begin{align*}
f_k(Z^{d_1};\lambda_k(\boldsymbol{x},\boldsymbol{y})) & = Z^{d_1}, \\
f_k(Z^{d_2};\lambda_k(\boldsymbol{x},\boldsymbol{y},Z^{d_1})) & = Z^{d_2} \odot \exp(\boldsymbol{s}_k) + \boldsymbol{t}_k,
\end{align*}
\noindent where it applies identity transform to $Z^{d_1}$ and applies a scale-shift transform to $Z^{d_2}$ according to $\boldsymbol{\lambda}_k = \{ \boldsymbol{s}_k, \boldsymbol{t}_k\}$, which are conditioned on $Z^{d_1}$, $\boldsymbol{x}$ and $\boldsymbol{y}$. CL is less expressive than PF and SF, but both sampling and computing the probability of arbitrary samples are easier. In practice, we follow \cite{realnvp45819} to switch $Z^{d_1}$ and $Z^{d_2}$ alternately for subsequent flows. 

As we adopt the amortized inference strategy, the parameters of these NFs are data-dependent. In our model, they are the output of 1-layer linear map with inputs that depend on $\boldsymbol{x}$ and $\boldsymbol{y}$. Also, as the introduction of normalizing flows no longer offers a simple closed-form solution, we modify the KL term in Eq.~\ref{eq_elbo1} into:
\begin{align*}
\mathbb{E}_{q_\lambda(Z \mid \boldsymbol{x}, \boldsymbol{y})} \left[ \log q_\lambda(Z \mid \boldsymbol{x}, \boldsymbol{y}) - \log p_\psi(Z \mid \boldsymbol{x}) \right] 
\end{align*}
where we estimate the expectation w.r.t.\ $q(Z_K{\mid}\boldsymbol{x};\boldsymbol{\lambda})$ via $L$ Monte-Carlo samples. We found that $L=1$ is sufficient, similar to \cite{zhang-etal-2016-variational-neural}.  To address variable-length inputs, we use the average of the embeddings of the source and target tokens via a mean-pooling layer, i.e., $\text{meanpool}(\boldsymbol{x})$ and $\text{meanpool}(\boldsymbol{y})$ respectively.
\\\\
\noindent \textbf{Transformer-based Generator}\quad We incorporate the latent code to the Transformer model by mixing the code into the output of the Transformer decoder's last layer ($h_j$) as follows:
\begin{align}
g_j &= \delta([h_j;Z]), \quad 
h_j &= (1-g_j) * h_j + g_j * Z \nonumber
\end{align}
where $g_j$ controls the latent code's contribution, and $\delta(\cdot)$ is the sigmoid function. In the case of the dimension of the latent code ($D$) doesn't match the dimension of $h_j$, we apply a linear projection layer. Our preliminary experiments suggest that Transformer is less likely to ignore the latent code in this approach compared to other approaches we explored, e.g., incorporating the latent code as the first generated token as used in \cite{zhang-etal-2016-variational-neural}.
\\\\
\noindent \textbf{Prediction} Ultimately, we search for the most probable translation $(\boldsymbol{\hat{y}})$ given a source sentence $(\boldsymbol{x})$ through the evidence lower bound. However, sampling latent codes from the posterior distribution is not straightforward, since the posterior is conditioned on the sentence being predicted. \newcite{zhang-etal-2016-variational-neural} suggests taking the prior's mean as the latent code. Unfortunately, as our prior is a  Gaussian distribution, this strategy can diminish the benefit of employing normalizing flows posterior. 

\newcite{autoencodingvnmt} explore two strategies, namely restricting the conditioning of the posterior to $\boldsymbol{x}$ alone (dropping $\boldsymbol{y}$) and introducing an auxiliary distribution, $r(Z{\mid}\boldsymbol{x})$, from which the latent codes are drawn. They found that the former is more accurate with the benefit of being simpler. This is confirmed by our preliminary experiments. We opt to adopt this strategy and use the mean of the posterior as the latent code at prediction time. 
\\\\
\noindent \textbf{Mitigating Posterior Collapse}\quad\quad As reported by previous work, VNMT models are prone to posterior collapse, where the training fails to learn informative latent code as indicated by the value of KL term that vanishes to 0. This phenomenon is often attributed to the strong generator \cite{DBLP:conf/icml/AlemiPFDS018} employed by the models, in which case, the generator's internal cells carry sufficient information to generate the translation. Significant research effort has been spent to weaken the generator network. Mitigating posterior collapse is crucial for our VNMT model as we employ the Transformer, an even stronger generator that comes with more direct connections between source and target sentences \cite{bahuleyan-etal-2018-variational}. 

To remedy these issues, we adopt the $\beta_C$-VAE \cite{prokhorov-etal-2019-importance} and compute the following modified KL term: $\beta \left| KL - C \right|$ where $\beta$ is the scaling factor while $C$ is a rate to control the KL magnitude. When $C>0$, the models are discouraged from ignoring the latent code. In our experiments, we set $C = 0.1$ and $\beta=1$. Additionally, we apply the standard practice of word dropping in our experiments. 
\\\\
\noindent \textbf{Related Work}\quad\quad VNMT comes in two flavors. The first variant models the conditional probability akin to a translation model, while the second one models the joint probability of the source and target sentences. Our model adopts the first variant similar to \cite{zhang-etal-2016-variational-neural,SuWXLHZ18,DBLP:journals/corr/abs-1812-04405}, while \cite{autoencodingvnmt,NIPS2018_7409} adopt the second variant.  The majority of VNMT models employ RNN-based generators and assume isotropic Gaussian distribution, except for \cite{mccarthy2019improved} and \cite{nfnmt}. The former employs the Transformer architecture but assumes a Gaussian posterior, while the latter employs the normalizing flows posterior (particularly planar flows) but uses an RNN-based generator. We combine more sophisticated normalizing flows and the more powerful Transformer architecture to produce state-of-the-art results. 

\section{Experimental Results}

\noindent\textbf{Experimental Setup}\quad We integrate our proposal into the Fairseq toolkit \cite{ott2019fairseq,gehring2016convenc,gehring2017convs2s}. We report results on the IWSLT’14 German-English (De-En) and the WMT'18 English-German (En-De) tasks. For IWSLT'14, we replicate \newcite{wu2018pay,edunov2018backtranslation}'s setup with 160K training sentences and a 10K joint BPE vocabulary, while for WMT'18, we replicate \newcite{edunov2018backtranslation}'s setup with 5.2M training sentences and a 32K joint BPE vocabulary.  For WMT experiments, we report the accuracy using detokenized SacreBLEU \cite{post-2018-call} to facilitate fair comparison with other published results. Note that tokenized BLEU score is often higher depending on the tokenizer, thus not comparable.  We apply KL annealing schedule and token dropout similar to \cite{bowman-etal-2016-generating}, where we set the KL annealing to 80K updates and  drop out 20\% target tokens in the IWSLT and 10\% in the WMT experiments.

The encoder and decoder of our Transformer generator have 6 blocks each. The number of attention heads, embedding dimension, and inner-layer dimensions are 4, 512, 1024 for IWSLT; and 16, 1024, 4096 for WMT. The WMT setup is often referred to as the \textit{Transformer Big}. To our knowledge, these architectures represent the best configurations for our tasks. We set the latent dimension to $D=128$, which is projected using a 1-layer linear map to the embedding space. We report decoding results with beam=5. For WMT experiments, we set the length penalty to 0.6. For all experiments with NF-based posterior, we employ flows of length 4, following the results of our pilot study.
\\\\
\noindent\textbf{In-Domain Results}\quad We present our IWSLT results in rows 1 to 6 of Table~\ref{t_result}. The accuracy of the baseline Transformer model is reported in row (1), which matches the number reported by \newcite{wu2018pay}.  In row (2), we report a static $Z$ experiment, where $Z=\text{meanpool}(\boldsymbol{x})$. We design this experiment to isolate the benefits of token dropping and utilizing average source embedding as context.  As shown, the static $Z$ provides +0.8 BLEU point gain. In row (3), we report the accuracy of our VNMT baseline when the approximate posterior is a Gaussian, which is +1.3 BLEU point from baseline or +0.5 point from the static $Z$, suggesting the efficacy of latent-variable modeling. We then report the accuracy of different variants of our model in rows (4) to (6), where we replace the Gaussian posterior with a cascade of 4 PF, SF and CL, respectively. For SF, we report the result with $M=8$ orthogonal columns in row (5). As shown, these flows modestly add +0.2 to +0.3 points. It is worth noticing that the improvement introduces only around 5\% additional parameters. 

\begin{table}[th]
\begin{center}
\begin{tabular}{|r|l|r|l|r|}
\hline
 & System & \#params & BLEU \\
\hline\hline
1 & Transformer IWSLT & 42.9M & 34.5 \\
\hline
2 & + static $Z$ & 42.9M & 35.3 \\
3 & + $Z \sim$ Gaussian & 43.6M  & 35.8\\
4 & + $Z \sim$ 4 x PF & \textit{44.2M}  & \textit{36.1} \\
5 & + $Z \sim$ 4 x SF (M=8) & 45.9M & 36.0 \\
6 & + $Z \sim$ 4 x CL & \textit{44.3M} & \textit{36.1} \\
\hline
7 & (1) + distilled & 42.9M & 34.9 \\
8 & (6) + distilled & 44.3M & \textbf{36.6} \\
\hline\hline
9 & \cite{edunov2018backtranslation} & \- & 29.0 \\
10 & Transformer Big & 209.1M & 28.9 \\ 
\hline
11 & + static $Z$ & 209.1M & 29.0 \\ 
12 & + $Z\sim$ Gaussian & 210.5M & 29.1 \\
13 & + $Z\sim$ 4 x PF & 211.6M & 29.3 \\ 
14 & $+ Z\sim$ 4 x SF (M=8) & 215.3M & \textit{29.5} \\
15 & $+ Z\sim$ 4 x CL & 210.6M & 29.2 \\
\hline
16 & (10) + distilled & 209.1M & 29.2 \\
17 & (14) + distilled & 215.3M & \textbf{29.9} \\
\hline
\end{tabular}

\end{center}

\caption{The translation accuracy on the De-En IWSLT'14 task (rows 1-8), the En-De WMT'18 task (rows 10-17). Each task's best results in the in-domain setting are \textit{italicized}, while the results with added distilled data are in \textbf{bold}. \label{t_result}}
\end{table}

We report our WMT results that use the Transformer Big architecture in rows (10) to (15). For comparison, we quote the state-of-the-art result for this dataset from \newcite{edunov2018backtranslation} in row (9), where the SacreBLEU score is obtained from \newcite{githuburl}. As shown, our baseline result (row 10) is on par with the state-of-the-art result. The WMT results are consistent with the IWSLT experiments, where our models (rows 13-15) significantly outperform the baseline, even though they differ in terms of which normalizing flows perform the best. The gain over the VNMT baseline is slightly higher, perhaps because NF is more effective in larger datasets. In particular, we found that SF and PF perform better than CL, perhaps due to their simpler architecture, i.e., their posteriors are conditioned only on the source sentence, and their priors are uninformed Gaussian. Row (11) shows that the static $Z$'s gain is minimal. In row (14), our best VNMT outperforms the state-of-the-art Transformer Big model by +0.6 BLEU while adding only 3\% additional parameters.  
\\\\
\noindent\textbf{Simulated Bimodal Data} \quad We conjecture that the gain partly comes from NF's ability to capture non-Gaussian distribution. To investigate this, we artificially increase the modality of our training data, i.e., forcing all source sentences to have multiple translations. We perform the sequence-level knowledge distillation \cite{kim-rush-2016-sequence} with baseline systems as the teachers, creating additional data referred to as \textit{distilled} data. We then train systems on this augmented training data, i.e., original + distilled data. Rows (7) and (16) show that the baseline systems benefit from the distilled data. Rows (8) and (17) show that our VNMT models gain more benefit, resulting in +2.1 and +0.9 BLEU points over non-latent baselines on IWSLT and WMT tasks respectively. 
\\\\
\noindent\textbf{Simulated Out-of-Domain Condition}\quad We investigate whether the in-domain improvement carries to out-of-domain test sets. To simulate an out-of-domain condition, we utilize our existing setup where the domain of the De-En IWSLT task is TED talks while the domain of the En-De WMT task is news articles. In particular, we invert the IWSLT De-En test set, and decode the English sentences using our baseline and best WMT En-De systems of rows (10) and (14). For this inverted set, the accuracy of our baseline system is 27.9, while the accuracy of our best system is 28.8, which is +0.9 points better. For reference, the accuracy of the Gaussian system in row (11) is 28.2 BLEU. While more rigorous out-of-domain experiments are needed, this result gives a strong indication that our model is relatively robust for this out-of-domain test set.
\\\\
\begin{table}[th]
\footnotesize
\begin{tabular}{p{15mm}|p{53mm}}
\hline
Example 1 Source & In \textit{\textcolor{blue}{her}} book , \textit{\textcolor{blue}{the interior decorator}} presents 17 housing models for independent living in old age .\\
\hline
Reference & In \textit{\textcolor{orange(ryb)}{ihrem}} Buch stellt \textit{\textcolor{orange(ryb)}{die Innenarchitektin}} 17 Wohnmodelle für ein selbstbestimmtes Wohnen im Alter vor .\\ 
\hline
Non-latent Baseline & In \textit{\textcolor{orange(ryb)}{ihrem}} Buch präsentiert \textit{\textcolor{red}{der Innenarchitekt}} 17 Wohnmodelle für ein unabhängiges Leben im Alter .\\
\hline
VNMT-G & In \textit{\textcolor{orange(ryb)}{ihrem}} Buch stellt die \textit{\textcolor{red}{der Innenarchitekt}} 17 Wohnmodelle für ein selbstbestimmtes Wohnen im Alter vor .\\
\hline
VNMT-NF &  In \textit{\textcolor{orange(ryb)}{ihrem}} Buch präsentiert \textit{\textcolor{orange(ryb)}{die Innendekoratorin}} 17 Wohnmodelle für ein unabhängiges Leben im Alter .\\

\hline\hline
Example 2 Source & Even though \textit{\textcolor{blue}{she}} earns S $ 3,000 ( $ 2,400 ) a month \textit{\textcolor{blue}{as an administrator}} and her husband works as well , the monthly family income is insufficient , she says .\\
\hline
Reference & Obwohl \textit{\textcolor{orange(ryb)}{sie}}  jeden Monat 3.000 Singapur-Dollar (ca 1.730 Euro ) \textit{\textcolor{orange(ryb)}{als Verwaltungsmitarbeiterin}} verdiene --truncated--\\ 
\hline
Non-latent Baseline & Obwohl \textcolor{ao(english)}{sie} pro Monat 3.000 S \$ ( 2.400 \$ ) \textcolor{red}{als Verwalter} verdient und auch ihr Mann arbeitet , ist das --truncated--\\
\hline
VNMT-G & Obwohl \textit{\textcolor{orange(ryb)}{sie}} jeden Monat 3.000 Singapur - Dollar ( ca 1.730 Euro ) \textit{\textcolor{orange(ryb)}{als Verwaltungsmitarbeiterin}} --truncated--\\
\hline
VNMT-NF &  Obwohl \textit{\textcolor{orange(ryb)}{sie}} S \$ 3.000 ( \$ 2.400 ) pro Monat \textit{\textcolor{orange(ryb)}{als Administratorin}} verdient und ihr Mann auch --trunctated--\\
\hline
\end{tabular}
\caption{Translation examples with different gender consistency. Inconsistent, consistent translations and source words are in \textit{\textcolor{red}{red}},  \textit{\textcolor{orange(ryb)}{orange}}, \textit{\textcolor{blue}{blue}} respectively. }\label{t_example}
\end{table}%
\noindent\textbf{Translation Analysis} To better understand the effect of normalizing flows, we manually inspect our WMT outputs and showcase a few examples in Table~\ref{t_example}. We compare the outputs of our best model that employs normalizing flows (VNMT-NF, row 14) with the baseline non-latent Transformer (row 10) and the baseline VNMT that employs Gaussian posterior (VNMT-G, row 12). 

As shown, our VNMT model consistently improves upon gender consistency. In example 1, the translation of \textit{\textcolor{blue}{the interior decorator}} depends on the gender of its cataphora (\textit{\textcolor{blue}{her}}), which is feminine. While all systems translate the cataphora correctly to \textit{\textcolor{orange(ryb)}{ihrem}}, the baseline and VNMT-G translate the phrase to its masculine form. In contrast, the translation of our VNMT-NF produces the feminine translation, respecting the gender agreement.  In example 2, only VNMT-NF and VNMT-G produce gender consistent translations. 

\section{Discussions and Conclusions}

We present a Variational NMT model that outperforms a strong state-of-the-art non-latent NMT model. We show that the gain modestly comes from the introduction of a family of flexible distribution based on normalizing flows.  We also demonstrate the robustness of our proposed model in an increased multimodality condition and on a simulated out-of-domain test set. 

We plan to conduct a more in-depth investigation into actual multimodality condition with high-coverage sets of plausible translations. We conjecture that conditioning the posterior on the target sentences would be more beneficial. Also, we plan to consider more structured latent variables beyond modeling the sentence-level variation as well as to apply our VNMT model to more language pairs.

\bibliography{anthology,acl2020}
\bibliographystyle{acl_natbib}

\appendix

\section{Word dropout}

We investigate the effect of different dropout rate and summarize the results in Table~\ref{t_dropout}. In particular, we take the VNMT baseline with Gaussian latent variable for IWSLT (row 3 in Table~\ref{t_result}) and for WMT (row 12 in Table~\ref{t_result}). As shown, word dropout is important for both setup but it is more so for IWSLT. It seems that tasks with low resources benefit more from word dropout. We also observe that above certain rate, word dropout hurts the performance.

\begin{center}
\begin{table}[h!]
\begin{tabular}{|l|r|r|r|r|}
\hline
Dropout rate & 0.0 & 0.1 & 0.2 & 0.3 \\
\hline
IWSLT & 34.4 & 35.7 & \textbf{35.8} & 35.6 \\
WMT & 29.0 & \textbf{29.1} & 28.8 & 28.7 \\
\hline
\end{tabular}
\caption{Results of different dropout rate for IWSLT and WMT setup. The best results are in \textbf{bold}.}\label{t_dropout}
\end{table}
\end{center}

\section{Latent Dimension}
We report the results of varying the dimension of latent variable ($D$) in Table~\ref{t_latdim}. For this study, we use the VNMT baseline with Gaussian latent variable in IWSLT condition  (row 3 in Table~\ref{t_result}) . Our experiments suggest that the latent dimension between 64 and 128 is optimal. The same conclusion holds for the WMT condition.
\begin{center}
\begin{table}[h!]
\setlength\tabcolsep{5pt}
\begin{tabular}{|l|r|r|r|r|r|r|r|r|}
\hline
$D$ &  8 & 16 & 32 & 64 & 128 & 256  \\
\hline
BLEU & 35.6 & 35.5 & 35.4 & 35.7& \textbf{35.8} & 35.4\\
\hline
\end{tabular}
\caption{Results of different dropout rate for IWSLT. The best results are in \textbf{bold}.}\label{t_latdim}
\end{table}
\end{center}

\section{Normalizing Flow Configuration}

In the Experimental Results section, we report the accuracy for our models with 4 flows. In Table~\ref{t_nf}, we conduct experiments varying the number of flows for the IWSLT condition. Our baseline (num flows=0) is an NMT model with word dropout, which performs on par with the static $Z$ experiment reported in Table~\ref{t_result}'s row 3. These results suggest that increasing the number of flows improves accuracy, but the gain diminishes after 4 flows. The results are consistent for all normalizing flows that we considered. We also conduct experiments with employing more flows, but unfortunately, we observe either unstable training or lower accuracy.

\begin{table}[h!]
\begin{center}
\begin{tabular}{|c|c|c|c|c|}
\hline
Num & \multirow{2}{*}{PF} & SF & \multirow{2}{*}{CL} \\
Flows & & ($M$=8)& \\
\hline
\hline
0 & \multicolumn{3}{c|}{35.3} \\ 
\hline
1  & 35.8 &  35.6 &  35.8 \\
2 & 35.7 & 35.5 & 35.8 \\
3  & 36.0 & 35.9 &  35.7 \\
4 &  \textbf{36.1} &  36.0 & \textbf{36.1} \\
5 &  35.9 &  \textbf{36.1} & 35.9 \\
6  & 35.8 &  36.0 &  35.9 \\
\hline
\end{tabular}
\caption{Translation accuracy of VNMT models employing various number of flows in the IWSLT condition. The best results are in \textbf{bold}.}\label{t_nf}
\end{center}
\end{table}

In Table~\ref{t_ortho}, we conduct experiments varying the number of orthogonal columns ($M$) in our Sylvester normalizing flows (SF) experiments. As shown, increasing $M$ improves the accuracy up to $M=24$. We see no additional gain from employing more additional orthogonal columns beyond 24. In Table~\ref{t_result}, we report $M=8$, because it introduces the least number of additional parameters.

\begin{center}
\begin{table}[h!]
\setlength\tabcolsep{5pt}
\begin{tabular}{|l|r|r|r|r|r|r|}
\hline
$M$ & 2 & 4 & 8 & 16 & 24 & 32   \\
\hline
BLEU & 35.7 & 35.5 & 36.0 & 36.0 & \textbf{36.2} & 35.9\\
\hline
\end{tabular}
\caption{Results of different number of orthogonal columns for SF. The best results are in \textbf{bold}.}\label{t_ortho}
\end{table}
\end{center}

\end{document}